\documentclass[11pt,a4paper]{article}
\usepackage[hyperref]{emnlp2018}
\usepackage{times}
\usepackage{latexsym}
\usepackage{graphicx}
\usepackage{CJKutf8}
\usepackage{amsmath}
\usepackage{amsfonts}

\usepackage{url}

\aclfinalcopy 

\newcommand\confname{EMNLP 2018}

\title{Data Augmentation for Neural Online Chat Response Selection}

\author{Wenchao Du\\
  Language Technology Institute \\
  Carnegie Mellon University \\
  Pittsburgh, PA 15213 \\
  {\tt wenchaod@cs.cmu.edu} \\
  \And
  Alan W Black \\
  Language Technology Institute \\
  Carnegie Mellon University \\
  Pittsburgh, PA 15213 \\
  {\tt awb@cs.cmu.edu} \\
}

\date{}

\begin{document}
\maketitle
\begin{abstract}
Data augmentation seeks to manipulate the available data for training to improve the generalization ability of models. We investigate two data augmentation proxies, permutation and flipping, for neural dialog response selection task on various models over multiple datasets, including both Chinese and English languages. Different from standard data augmentation techniques, our method combines the original and synthesized data for prediction. Empirical results show that our approach can gain 1 to 3 recall-at-1 points over baseline models in both full-scale and small-scale settings.
\end{abstract}

\section{Introduction}
Building machines that are capable of conversing like humans is one of the primary goals of artificial intelligence. Extensive manual labor is typically required by traditional rule-based systems, limiting the scalability of such systems across multiple domains. With the success of machine learning, the quest of building data-driven dialog systems has come into focus over the past few years \cite{ritter2011data}. Existing approaches in this area can be categorized into generation-based methods and retrieval-based methods. While generation-based methods are still far from reliably generating informative responses, retrieval-based methods have the advantage of fluency and groundedness, since they select responses from existing data. We concentrate on retrieval-based methods in this paper, though we believe the proposed techniques could also improve generation-based models.

While current state-of-the-art results for dialog models are achieved by deep learning approaches, the performance of neural models largely depends on the amount of training data. However, acquiring conversational data can be difficult at times. On the other hand, even with thousands of data points, it is unclear whether these models can optimally benefit from them. Therefore, data augmentation and its efficient use becomes an important problem. Our main contribution is that we investigated new ways to manipulate chat data and neural model architectures to improve performance. To our knowledge, we are the first to evaluate data augmentation on different types of neural conversation models over multiple domains and languages.

\section{Data Augmentation}
Recent studies \cite{adi2016fine, khandelwal2018sharp} have shown that recurrent neural networks (RNN), especially long-short term memory networks (LSTM), are sensitive to word order when encoding contextual information. However, for the response selection task, it is so far unclear to what extent word order is important. This problem is perplexed by the following language phenomena we observed from existing chat data:
\begin{enumerate}
\item {Broken continuity. Simultaneous conversations happen in multi-party dialogs \cite{elsner2008you} very often, resulting in some utterances not responding to their immediately preceding ones. Even in conversations between only two people, continuity may still break due to one person switch topic before the other responds. See \hyperref[tab:first]{Table \ref*{tab:first}} for examples.}

\begin{table}
\begin{CJK*}{UTF8}{gbsn}
Example 1:
\begin{center}
\begin{tabular}{|l|l|}
\hline
Old & I dont run graphical ubuntu, \\
    & I run ubuntu server. \\
\hline
Kuja & Haha sucker. \\
\hline
Taru & ? \\
\hline
Burner & you can use "ps ax" and "kill (PID\#)" \\
\hline
Kuja & Anyways, you made the changes \\
     & right? \\
\hline
\end{tabular}
\end{center}
\vspace{1em}
Example 2:
\begin{center}
\begin{tabular}{|l|l|}
\hline
Customer & 在(there) 吗(?) \\
\hline
Customer & 看看(look at) 此(this) 款(one) \\
\hline
Agent & 在的(I'm here) 亲(dear) \\
\hline
Agent & 亲(dear)， 请(please) 发(send) \\
      & 链接(link) \\
\hline
\end{tabular}
\caption{Example chat snippets for broken continuity. The first example is from \cite{lowe2015ubuntu}. Burner's message is responding to Old, and Kuja's last message is replying to Taru. The second example is from Taobao, where the third message is responding to the first message, and the fourth message to the second message.
  }
\label{tab:first}
\end{center}
\end{CJK*}
\end{table}

\begin{table}
\begin{CJK*}{UTF8}{gbsn}
Example 1:
\begin{center}
\begin{tabular}{|l|l|}
\hline
Customer A & 这(this) 款(one) 我(I) 穿(wear) \\
           & 什么(what) 码(size) \\
\hline
Customer A & 160高(tall)，107 斤(0.5kg) \\
           & 重(heavy) \\
\hline
Agent & 亲(dear) 如果(if) 喜欢(like) \\
      & 宽松(loose) 点 的 就(then) \\
      & 可以(can) 选(choose) L 哦 \\
\hline
\end{tabular}
\end{center}
\vspace{1em}
Example 2:
\begin{center}
\begin{tabular}{|l|l|}
\hline
Customer B & 158cm \\
\hline
Customer B & 63kg \\
\hline
Customer B & 穿(wear) 什么(what) 码(size) \\
           & 的 合适(fit) \\
\hline
Agent & 亲(dear) 根据(based on) \\
      & 亲的(your) 数据(data)， \\
      & 建议(suggest) 穿(wear) \\
      & L 码(size) \\
\hline
\end{tabular}
\end{center}
\caption{Example chat snippets for mixed turn-taking from Taobao. The question for recommendation and its relevant information (height and weight) can be communicated through different number of utterances in arbitrary order.
  }
\label{tab:second}
\end{CJK*}
\end{table}

\begin{table}
\begin{CJK*}{UTF8}{gbsn}
Example:
\begin{center}
\begin{tabular}{|l|l|}
\hline
Wizard & Sorry, I cannot find any trips \\
       & leaving from Gotham City. Could \\
       & you suggest another nearby \\
       & departure city? \\
\hline
Customer & Would any packages to Mos Eisley \\
         & be available, if I increase my \\
         & budget to \$2500? \\
\hline
Wizard & There are no trips available to \\
       & Mos Eisley. \\
\hline
\end{tabular}
\end{center}
\caption{Example chat snippets from Frames. The first message has two sentences. The second message is a conditional complex sentence.
  }
\label{tab:third}
\end{CJK*}
\end{table}

\begin{table}
\begin{CJK*}{UTF8}{gbsn}
Example 2 of Table 1 after Permutation:
\begin{center}
\begin{tabular}{|l|l|}
\hline
Customer & 在(there) 吗(?) \\
\hline
Agent & 在的(I'm here) 亲(dear) \\
\hline
Customer & 看看(look at) 此(this) 款(one) \\
\hline
Agent & 亲(dear)， 请(please) 发(send) \\
      & 链接(link) \\
\hline
\end{tabular}
\end{center}
\vspace{1em}
Example 1 of Table 2 after Permutation:
\begin{center}
\begin{tabular}{|l|l|}
\hline
Customer A & 160高(tall)，107 斤(0.5kg) \\
           & 重(heavy) \\
\hline
Customer A & 这(this) 款(one) 我(I) 穿(wear) \\
           & 什么(what) 码(size) \\
\hline
Agent & 亲(dear) 如果(if) 喜欢(like) \\
      & 宽松(loose) 点 的 就(then) \\
      & 可以(can) 选(choose) L 哦 \\
\hline
\end{tabular}
\end{center}
\vspace{1em}
Example of Table 3 after Flipping:
\begin{center}
\begin{tabular}{|l|l|}
\hline
Wizard & Could you suggest another nearby \\
       & departure city? Sorry, I cannot find \\
       & any trips leaving from Gotham City. \\
\hline
Customer & if I increase my budget to \$2500, \\
         & Would any packages to Mos Eisley \\
         & be available? \\
\hline
Wizard & There are no trips available to Mos \\
       & Eisley. \\
\hline
\end{tabular}
\end{center}
\caption{Results of proposed transformations on previous examples. In the first and second examples, the two messages right before the last agent's response are permuted. In the third example, the first message is flipped, splitting at the period; the second messages is separated at the comma and flipped.
  }
\label{tab:fourth}
\end{CJK*}
\end{table}

\item {Mixed turn-taking behavior. People can give multiple utterances before the other respond. Usually, these consecutive messages from same person form arguments that are in parallel (by 'argument' we mean text spans that form discourse relations with each other), and their orderings are not that important. We found this to be very common in online live chats. See \hyperref[tab:second]{Table \ref*{tab:second}} for examples.}
\item {Long utterances. Some utterances contain multiple sentences. Some are single compound sentence with multiple clauses. See \hyperref[tab:third]{Table \ref*{tab:third}} for examples.}
\end{enumerate}

To summarize, the critical information for responding, which can be either a single word, phrase, or a full sentence, may have varying relative positions in the context. Therefore, we hypothesize that there exist alternative orderings of utterances and intra-utterance arguments in chat data that can help selecting responses, given recurrent neural models' sensitivity to word order. In this paper, our main goal is to seek improvement by creating variations in the ordering of utterances and arguments. We aim for \textit{generic} methods, bypassing the need of discourse and syntactic parsing as an intermediate step. With the fact that online chats are typically noisy with spelling errors and ungrammaticality, a relative lack of precision may actually help. We therefore propose the following ways to manipulate chat data:

\textbf{Permutation} is simply reversing the order of any two messages in the context. This may help recover the continuity or create alternative ordering of parallel arguments.

\textbf{Flipping} breaks an utterance into two parts, and concatenate them in their reversed order. The break point is the punctuation that is closest to the middle of the utterance if there is any. Otherwise, we break the utterance at the middle.

As illustrated in \hyperref[tab:fourth]{Table \ref*{tab:fourth}}, the proposed transformations neither change the implication of the contexts nor the appropriateness of the responses.

\begin{table*}
\centering
\begin{tabular}{llllllll}
   & \textbf{Language} & \textbf{Medium} & \textbf{Style} & \textbf{Domain} & \textbf{Size (Train)} & \textbf{Vocabulary} \\
  \hline
    \textbf{Ubuntu} & English & Chat Room & Noisy & Task & 1M & 400k \\
  \hline
    \textbf{Taobao} & Chinese & Chat Room & Noisy & Task & 0.9M & 90k \\
    \hline
    \textbf{Douban} & Chinese & Web Forum & Noisy & Open & 1M & 300k \\
    \hline
    \textbf{Frames} & English & Chat Room & Controlled & Task & 11k
& 9k \\ \hline

\end{tabular}
\caption{Comparison of four dialog corpora
  }
\end{table*}

\section{Data}
We describe four datasets that we will be using to evaluate our proposed methods:

\textbf{Taobao} chat log was collected by a vendor of pajamas between 2013 and 2015. The conversations took place on Taobao, one of the largest Chinese e-commerce websites. The website allows two-way conversations between customers and agents in individual sessions. 

\textbf{Ubuntu} dialog corpus \cite{lowe2015ubuntu} is the first large dataset of online chats made available. It contains multi-party chat logs from Ubuntu chat room where people help each other to solve technical problems related to Ubuntu. 

\textbf{Douban} conversation corpus is a collection of web forum post discussions from Douban, a Chinese internet community \cite{wu2016sequential}. It covers a wide range of topics, hence open-domain in nature. 

\textbf{Frames} dataset was collected by \cite{asri2017frames} in wizard-of-oz setting. The chats are about booking flight. The wizard has access to database to answer domain-specific questions. Unlike the datasets mentioned above, the conversations of Frames are highly controlled so that the language is perfect and the chats have perfect turn exchanges.

\section{Methodology}
\subsection{Model Overview}
We first give a high level abstraction of the neural models we will be investigating. Given context and candidate responses, the models score each candidate and the one with the highest score is selected. The models are trained by maximizing the likelihood of labels. To build training data, one negative example is sampled from the corpus for each pair of context and true response. We group the models into the following two categories:

\textbf{Dual-Encoder Model (DE)} As first proposed in \cite{lowe2015ubuntu}, DE models encode context $m$ and response $r$ into $v(m) \in \mathbb{R}^l, v(r) \in \mathbb{R}^m$, respectively. Then
$$ P(r \mid m) = \sigma(v(m)^T M v(r)) $$
where $\sigma$ is the sigmoid function, $M \in \mathbb{R}^{l \times m}$. In this paper, response encoder is LSTM. We consider two choices of context encoder: one is word-level LSTM encoder only (LSTM-DE), which takes concatenated messages as input. The other one is hierarchical recurrent encoder (HRE-DE). For HRE, we encode each message with an LSTM word-level encoder, and then feed the last hidden states from the word-level encoder to an utterance-level encoder, which is also an LSTM. We concatenate the last hidden state of the utterance-level encoder to that of word-level encoder on concatenated messages as final context encoding. Note that HRE-DE is a simplified version of the model in \cite{zhou2016multi}.

\textbf{Sequential Matching Network (SMN)} Unlike DE models, SMN finds the affinity between context messages and responses as a first step \cite{wu2016sequential}. Given messages $m_k$ where $k=1,...,n$ and response $r$, SMN first extract feature $u(m_k, r) \in \mathbb{R}^p$ of how related the two utterances are, and then accumulate these features with an RNN:
$$v(m, r) = RNN(u(m_k, r)), k = 1,...,n $$
$$P(r \mid m) = \sigma(w^T v(m, r)) $$
where $v(m, r), w \in \mathbb{R}^q$.

\subsection{Combining Transformed Data}
Let $\pi_i$ be the applicable transformations including the identity. For context $m$ and response $r$, let $m^i = \pi_i(m)$, $r^j = \pi_j(r)$. For DE models, we use the same encoder for $m$, $r$ to encode $m^i$, $r^j$. Then we combine the encodings and predict by
$$ P(r \mid m) = \sigma( \sum_{i, j}  v(m^i)^T M_{ij} v(r^j)) $$
where $M_{ij} \in \mathbb{R}^{l \times m}$. Similarly, for SMN, the predicted score is 
$$P(r \mid m) = \sigma(\sum_{i,j} w^T_{i,j} v(m^i, r^j)) $$
where $w_{i,j} \in \mathbb{R}^q$. Please note that this score function allows augmentations to be done at test time for prediction. Additionally, we inject squared distance between the encodings of the original data and the transformed data in order to enforce models to learn similar representations for them. We are assuming that the transformation should not drastically change the meanings of contexts and responses even though they are not exactly label-preserving. Empirically we found adding this regularization term actually helps. The training loss for DE models becomes
\begin{align*}
\sum_{(m, r)} & ( -\log P(r \mid m) + t(\sum_i \| v(m^i) - v(m) \|^2 \\
& + \sum_j  \| v(r^j) - v(r) \|^2)
\end{align*}
and the one for SMN becomes
$$ \sum_{(m, r)} ( -\log P(r \mid m) + t(\sum_{i,j} \| v(m^i, r^j) - v(m, j) \|^2) $$
where $t$ is a hyper-parameter. We tuned it on the validation set in $[0.01, 0.1]$.

\begin{table*}[h!t]
\centering
\resizebox{2\columnwidth}{0.135\textheight} {
\begin{tabular}{|c||c|c||c|c||c|c||c|}
\hline
   & \multicolumn{2}{c||}{\textbf{Ubuntu}} & \multicolumn{2}{c||}{\textbf{Taobao}} & \multicolumn{2}{c||}{\textbf{Douban}} & \textbf{Frames} \\
  \hline
  & 100\% & 1\% & 100\% & 1\% & 100\% & 1\% & 100\% \\
  \hline
    LSTM-DE & 0.6546 & 0.3470 & 0.8446 & 0.4862 & 0.6193 & 0.3301 & 0.3941  \\

      + permutation 1 & 0.6773 & 0.3723 & \textbf{0.8685} & 0.5037 & \textbf{0.6402} & \textbf{0.3503} & 0.3973 \\

      + permutation 2 & \textbf{0.6854} & 0.3685 & \textbf{0.8693} & 0.5071 & \textbf{0.6469} & 0.3444 & 0.4122 \\

      + flipping & \textbf{0.6853} & \textbf{0.3778} & \textbf{0.8669} & \textbf{0.5201} & \textbf{0.6430} & 0.3369 & \textbf{0.4209} \\
\hline 
    HRE-DE & 0.6729 & 0.3654 & 0.8728 & 0.5085 & 0.6443 & 0.3350 & 0.4436 \\
  
      + permutation 1 & 0.6817 & 0.3650 & 0.8732 & 0.5053 & 0.6401 & 0.3423 & 0.4339 \\
 
      + permutation 2 & 0.6786  & \textbf{0.3713} & 0.8787 & \textbf{0.5207} & 0.6430 & 0.3395 & \textbf{0.4518} \\

        + flipping & \textbf{0.6920} & 0.3688 & \textbf{0.8828} & 0.5147 & \textbf{0.6542} & \textbf{0.3523} & \textbf{0.4564} \\
  \hline 
    SMN & 0.7050 & 0.4771 & 0.8194 & 0.5312 & 0.6700 & 0.4662 & 0.4055 \\
 
      + permutation 1 & 0.7066 & 0.4749 & 0.8171 & 0.5302 & 0.6747 & 0.4669 & 0.4023 \\

      + flipping & \textbf{0.7156} & \textbf{0.4893} & \textbf{0.8231} & \textbf{0.5387} & \textbf{0.6800} & \textbf{0.4876} & \textbf{0.4116} \\
  \hline
\end{tabular}
}
\caption{Numbers on recall-at-1. Best results for each dataset and each model are highlighted.}
\label{tab:sixth}
\end{table*}

\section{Experiments}
\subsection{Setup}
We evaluate our method on the datasets mentioned in Section 3. For the Ubuntu dataset, we use the version shared by \cite{xu2016incorporating}. For Douban, we discard the test set provided by the authors since the responses are not from the same domain, and re-split training set. Negative responses are randomly sampled. For Frames, we select negative responses from those that have different slot types and values from true responses. We also conduct an experiment with smaller amount of training data on the three large datasets, Ubuntu, Douban, and Taobao, in which $1\%$ of the training set are randomly selected for training. Following \cite{lowe2015ubuntu}, we evaluate the model performance with recall-at-1, following previous work.

We experiment with two types of permutation: the first one is permuting the last and the penultimate message in contexts, and the second one is permuting the penultimate with the third to last message. We only do the first type of permutation for SMN since SMN seems to be insensitive to permutation. We flip all messages in contexts and responses for SMN, and only flip context messages for DE models.

\subsection{Training}
We initialize word embeddings using the results of word2vec \cite{mikolov2013distributed} trained on the whole corpus. The size of word embeddings is 300 for LSTM-DE and HRE-DE, and 200 for SMN. For LSTM-DE and HRE-DE, each LSTM layer has hidden size of 300. We use the same hyper-parameters for SMN as in \cite{wu2016sequential}. All models are trained with Adam optimizer with learning rate of $0.001$. We use early stopping to choose parameters. For experiments on small training sets (including Frames), we additionally apply dropout \cite{srivastava2014dropout} with rate $0.5$ to all recurrent layers. As a side note, we find that dropout does not affect the result in any significant way under full-scale setting.

\subsection{Main Results}
\hyperref[tab:sixth]{Table \ref*{tab:sixth}} shows the performance of LSTM-DE, HRE-DE, and SMN on 4 different datasets under different types of augmentation. For each full-scale dataset, nearly all models gain around 1 to 3 points with one of the proposed data augmentation methods. Permutation works best for LSTM-DE, less so for HRE-DE, and has almost no effect on SMN. This is probably because HRE-DE and SMN have an utterance-level recurrent component which makes them better at capturing long range dependencies. Permutation 1 does not improve on Frames dataset for any model. This might be that Frames has perfect turn-taking, and wizards' responses are mostly addressing their immediately preceding messages, so moving away the last message in context does not help. In small-scale setting, LSTM-DE with data augmentation outperforms HRE-DE on some of the datasets. SMN gains even more with flipping than in full-scale setting.

\section{Related Work}
Data augmentation has been widely adopted in computer vision and speech recognition \cite{krizhevsky2012imagenet, ko2015audio}. In image processing, label-preserving transformations such as tilting and flipping are used, but in NLP, finding such transformations that exactly preserve meanings is difficult. Language data is discrete in nature, and minor perturbation may change the meaning. Most commonly used techniques include word substitution \cite{fadaee2017data} and paraphrasing \cite{dong2017learning}. These methods may require heavy external resources, which can be difficult to apply across multiple languages and domains.

Recently, there has been a surging interest in adversarial training \cite{goodfellow2014explaining}. For text data, one class of methods generate adversarial examples by moving word embeddings along the opposite direction of the gradient of loss functions \cite{wu2017adversarial,yasunaga2017robust}, hence small perturbation in the continuous space of word vectors. Another class of methods aim to create genuinely new examples. \cite{li2017robust} adds syntactic and semantic variations to training data based on grammar rules and thesaurus. \cite{xie2017data} add noises to data by blanking out or substituting words for language modeling. \cite{yang2017semi} adopt a seq2seq model \cite{sutskever2014sequence} to generate questions based on paragraphs and answers into their generative adversarial framework. One main difference between these methods and our approach is that, while adversarial training only manipulates training data, we in addition apply transformations to data \textit{at test time} to help prediction. This is closer to \cite{dong2017learning} in spirit.

\section{Conclusion}

We proposed a general method to improve dialog response selection through manipulating existing data that can be applied to different models. Our results show that for both open-domain and task-oriented dialogues, and for both English and Chinese languages, at least one of the proposed augmentation methods is effective, and the chance that they hurt is rare. We have deliberately chosen a diverse set of domains and models to test this on to try to understand the contribution of data augmentation. Thus even when working on new datasets, and new models, it seems data augmentation is still a valuable addition that will likely improve results. Being more specific about when augmentation works is harder. One future research direction would be to apply data transformation situationally based on the discourse structure of dialogs. In our experiments, we tried combining permutation and flipping but found no advantage over using only one type of transformation. We believe a more sophisticated method of combination could further improve the results, and leave it to future work.

\bibliography{emnlp2018}
\bibliographystyle{acl_natbib_nourl}

\end{document}